\documentclass[letterpaper, 10 pt, conference]{ieeeconf}  

\IEEEoverridecommandlockouts                              

\usepackage{graphicx}
\usepackage{amsmath,amssymb,amsfonts}
\usepackage[caption=false]{subfig}
\usepackage[ruled,linesnumbered, noend]{algorithm2e}
\usepackage{dirtytalk}
\usepackage{hyperref}
\usepackage{gensymb}
\usepackage{tabularx}
\usepackage{multirow}

\makeatletter
\newcommand{\removelatexerror}{\let\@latex@error\@gobble}
\makeatother
\SetKwComment{Comment}{$\triangleright$\ }{}

\title{\LARGE \bf
Autonomous UAV Exploration of Dynamic Environments via Incremental Sampling and Probabilistic Roadmap
}

\author{Zhefan Xu, Di Deng, and Kenji Shimada 
\thanks{*This work was partially supported by TOPRISE Co, Ltd.}
\thanks{Zhefan Xu, Di Deng, and Kenji Shimada are with Department of Mechanical Engineering, Carnegie Mellon University, 5000 Forbes Ave, Pittsburgh, PA, 15213, USA.
        {\tt\small zhefanx@andrew.cmu.edu}}%
}

\begin{document}

\maketitle
\thispagestyle{empty}
\pagestyle{empty}

\begin{abstract}

Autonomous exploration requires robots to generate informative trajectories iteratively. Although sampling-based methods are highly efficient in unmanned aerial vehicle exploration, many of these methods do not effectively utilize the sampled information from the previous planning iterations, leading to redundant computation and longer exploration time. Also, few have explicitly shown their exploration ability in dynamic environments even though they can run real-time. To overcome these limitations, we propose a novel dynamic exploration planner (DEP) for exploring unknown environments using incremental sampling and Probabilistic Roadmap (PRM). In our sampling strategy, nodes are added incrementally and distributed evenly in the explored region, yielding the best viewpoints. To further shortening exploration time and ensuring safety, our planner optimizes paths locally and refine them based on the Euclidean Signed Distance Function (ESDF) map. Meanwhile, as the multi-query planner, PRM allows the proposed planner to quickly search alternative paths to avoid dynamic obstacles for safe exploration. Simulation experiments show that our method safely explores dynamic environments and outperforms the benchmark planners in terms of exploration time, path length, and computational time. 

\end{abstract}

\section{INTRODUCTION}

The autonomous exploration technique can be used in many different industrial applications such as inspection, surveillance, rescue, and 3D reconstruction. In recent years, the robotics community has paid more attention to the usage of unmanned aerial vehicles (UAV) in these applications because of UAVs' low cost, agility, and flexibility. All these applications require UAVs to determine informative paths iteratively.

The online exploration problem can be viewed as how to determine a series of informative sensor positions \cite{1}. Early frontier exploration proves the success in gaining knowledge of the unknown region by visiting the border using 2D ground robots \cite{2}. Although later works \cite{3} \cite{4} extend the idea to adapt the aerial robots, this method still suffers from the expensive computation in high dimensional planning. In contrast, the sampling-based method is preferred for UAV exploration because of its high computational efficiency and the various reliable information gain formulations. Typically, these methods \cite{5} \cite{6} \cite{7} \cite{8} \cite{9} \cite{10} apply the cost-utility function to evaluate the exploration potentials of the sampling nodes and obtain paths based on the single-query planner such as the rapid-exploring random tree (RRT) \cite{11} or its variants. However, the limited number of random sampling in a single iteration cannot guarantee the comprehensive coverage of nodes in the mapped space, which degrades the selection of the optimal viewpoints. Also, the computation can be wasted by repeatedly evaluating the region already sampled in previous iterations. Since the exploration is an iterative and repeated process, a multi-query planner such as probabilistic roadmap (PRM) \cite{12} can be more suitable for reducing computational cost and determining the informative viewpoints based on the accumulated knowledge. Even though some tree-based method \cite{8} continuously adds nodes to the tree branches or reuses historical information \cite{6} to overcome these problems, they still lack the flexibility to quickly generate the suboptimal alternative paths to avoid dynamics obstacles.

\begin{figure}[t]
    \vspace{0.2cm}
    \centering
    \includegraphics[scale=0.335]{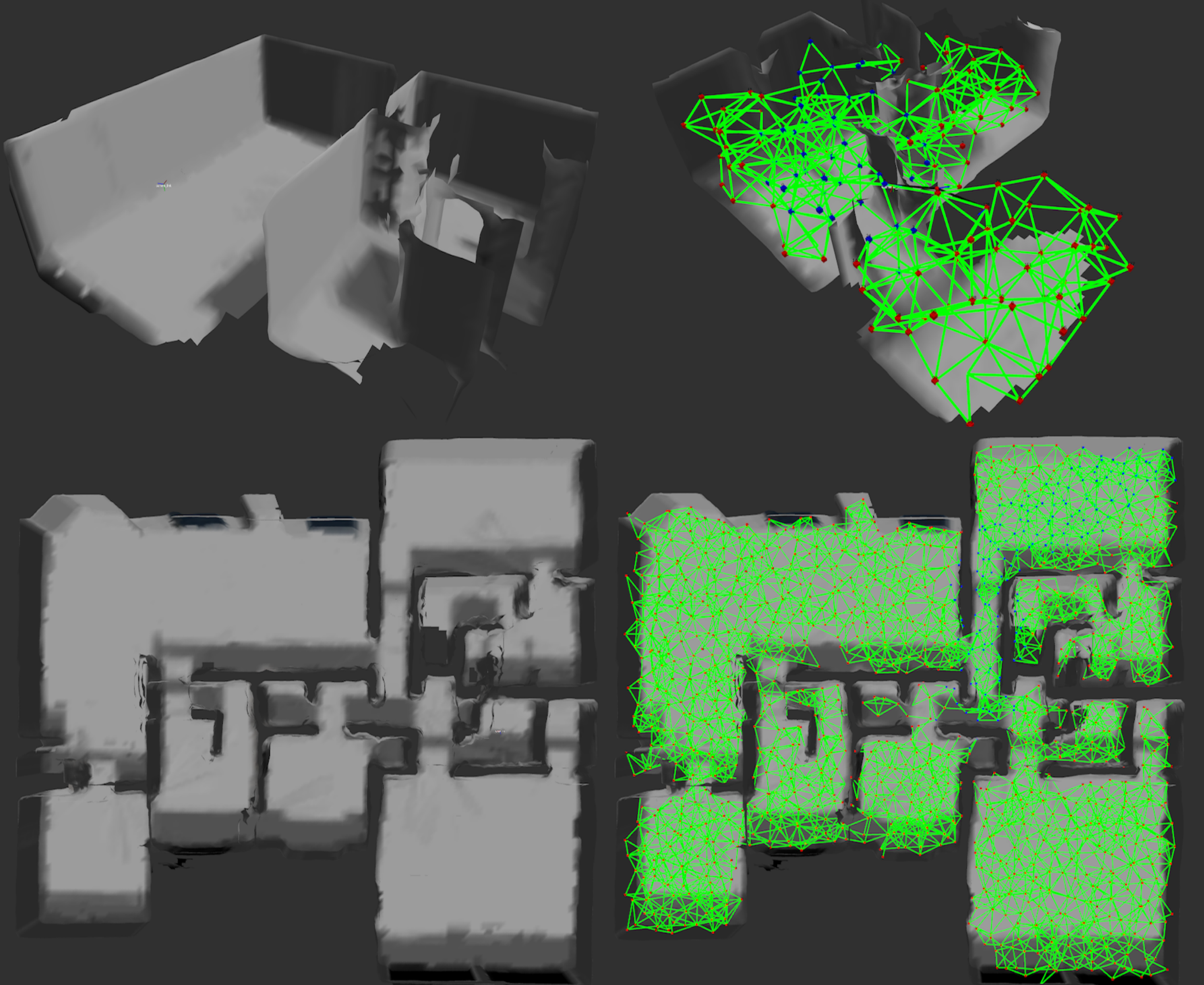}
    \caption{Visualization of an explored map (left) and incrementally constructed roadmap (right) for a large office environment. Green lines and red dots represent edges and nodes, respectively. Note that blue dots are the nodes whose exploration utilities (gains) need to be updated.}
    \label{office_vis}

\end{figure}

To address these issues, we propose a multi-query dynamic exploration planner (DEP) based on PRM. Unlike traditional sampling-based methods, our incremental sampling strategy effectively improves the mapped region's node coverage. As the nodes' utilities are stored and updated as the intrinsic attributes, trajectory generation can be performed quickly to avoid dynamic obstacles. The gain rate evaluation of multiple trajectory candidates avoids the planner's greedy behaviors. ESDF-based optimization is later applied to minimize the trajectory execution time and maintain the tolerant distance to obstacles. The results show that our method outperforms the benchmark algorithms by total exploration time, computational time, and path length. Fig. \ref{office_vis} shows the example of our planner running in a large office environment with roadmap visualization. 

The novelties and contributions of this work are:
\begin{itemize}
  \item \textbf{Incremental PRM for Exploration:} This work presents a new online exploration algorithm DEP\footnote{https://github.com/Zhefan-Xu/DEP} inspired by incremental PRM. Our roadmap evenly grows sampled nodes with gain evaluation in an incremental fashion. The highest utility trajectory is selected by our trajectory generation method for the robot to execute.
  \item \textbf{Safety in Dynamic Environments:} To the best of the authors' knowledge, our DEP planner is the first method to explicitly consider exploration safety in unknown environments with dynamic obstacles. Our proposed incremental PRM and trajectory optimization achieves faster replanning and safe obstacle distances.
  \item \textbf{High Exploration Efficiency:} We conduct thorough comparison experiments with classic \cite{2} and recent state-of-the-art \cite{5}\cite{7} planners. The results indicate that our planner has the highest efficiency in exploration.
\end{itemize}

\section{Related Work}
There are two major categories of autonomous exploration algorithms: the frontier-based method and the sampling-based method. Early frontier exploration proves high efficiency in 2D ground robots \cite{2}, and it is later extended to support fast-flying UAV exploration \cite{13} and 3D reconstruction \cite{14}. In contrast, instead of guiding the exploration by frontiers, the sampling-based method selects the best viewpoints using the information gain as the exploration heuristic \cite{1}\cite{15}\cite{16}. As the computation of the sampling-based method does not increase significantly with the robot dimension, it is preferred in the UAV application.  Other approach like the information-theoretic method provides a sophisticated way for information gain evaluation \cite{17}, and the data-driven approach \cite{18} has also been experimented with but are hard to be generalized for different environments. There are also recent research focusing on the heterogeneous robot exploration \cite{19}.

The sampling-based method is more popular in recent years for UAV exploration. Receding horizon next-best-view (RH-NBV) planner brings a reliable solution to robot exploration \cite{5}. By growing a tree from the robot's current position, the best branch with the highest information gain is selected, and the robot executes the first branch segment. There exist many recent works inspired by this approach. \cite{9} adopts a two-stage planner to optimize the saliency gain to consider the visual saliency of different objects in the environment. To improve the sampling efficiency and avoid the planner from getting trapped in the local minima, \cite{6} stores a historical graph from the previous samples for determining the exploration potentials. It also comes up with the orientation angle optimization for better gain estimation. Similarly,  \cite{7} combines RH-NBV with the frontier-based algorithm to prevent early termination in local minima in their autonomous exploration planner (AEP). The frontiers are cached for determining the exploration goals, and the cached nodes can help estimate the information gain by Gaussian process. In \cite{8}, their RRT*-based planner continuously grows and maintains the tree with rewiring for path refinement to avoid discarding the rest of the nodes not in the best branch. Their TSDF-based reconstruction gain results in the lowest 3D reconstruction errors compared to other sampling-based methods. For reducing the mapping and localization error, \cite{20}, \cite{21}, and \cite{22} consider the localization uncertainty in their planning steps. 

\section{Problem Description}
A bounded environment, $V_\text{b} \subset \mathbb{R}^\text{3}$, consists of the free space, $V_{\text{free}}$, and occupied space, $V_{\text{occ}}$. A UAV with a depth camera is used for exploration. Due to constraints on the robot size, environment geometry, and sensor range,  there exists some sensor-unreachable space, $V_{\text{ur}}$. Occupancy map $\mathcal{M}$ is divided into small voxels with resolution $r$. Initially, the whole environment is unknown except the nearby region, $V_{\text{mapped}}^{\text{init}}$, around the robot. There are both static obstacles, $O_{\text{static}}$, and dynamic obstacles, $O_{\text{dynamic}}$, in the environment. 

\it{Problem I: Autonomous Exploration:} \normalfont{ With the initial mapped space, $V_{\text{mapped}}^{\text{init}}$, the robot needs to generate a collision-free path, $\sigma$, consisting of the waypoints, $p$, from a set of valid configurations in map, $\mathcal{M}_{\text{valid}}$. By executing the path, robot can enlarge $V_{\text{mapped}}$ by sensor scanning.  The task is finished when the robot maps all the reachable space, $V_{\text{mapped}} = (V_{\text{free}} \cup V_{\text{occ}}) \setminus V_{\text{ur}}$.}

\it{Problem II: Dynamic Obstacle Avoidance}: \normalfont{During the path execution, the robot needs to constantly monitor unexpected dynamic obstacles, $O_{\text{dynamic}}^{\text{traj}}$, which are on the robot's executing path segment. Since the sensor range is limited, there is one extra constraint on the waypoints of the path: $p_{\text{next}}$ must be in the sensor range of $p_{\text{current}}$ . When $O_{\text{dynamic}}^{\text{traj}}$ is detected, the robot must replan in a reasonable time to avoid collisions.}

\section{Proposed Method}
\begin{figure}[t]
    \centering
    \includegraphics[scale=0.31]{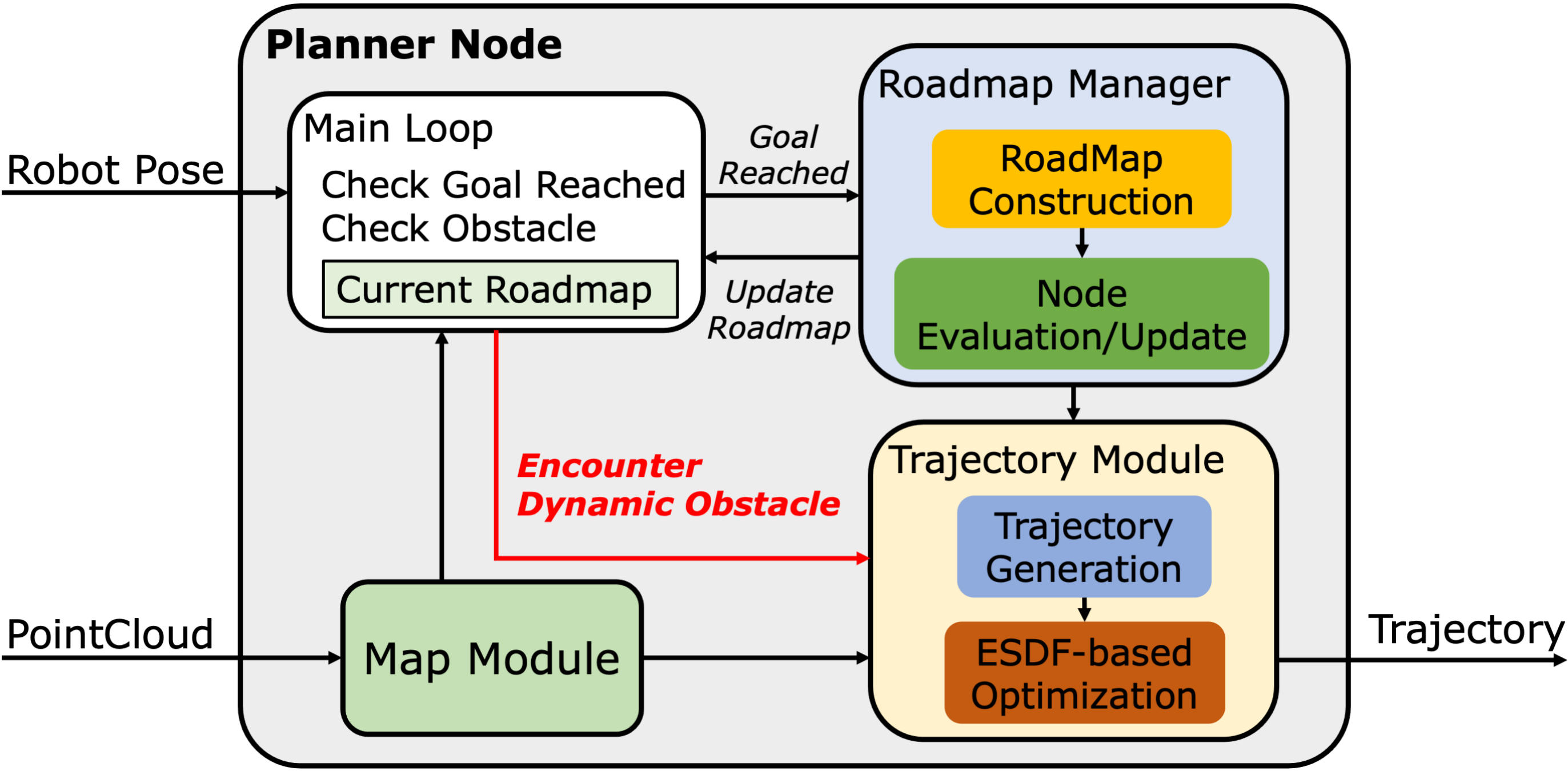}
    \caption{Planner overview. The roadmap is incrementally updated in each planning iteration for generating the trajectory. When dynamic obstacles are encountered, the current roadmap can be used to generate new trajectories.}
    \label{algorithm_flowchart}
\end{figure}

\begin{figure*}[t]
\centering
\subfloat[Robot completes the previous trajectory]{\includegraphics[width=.325\linewidth]{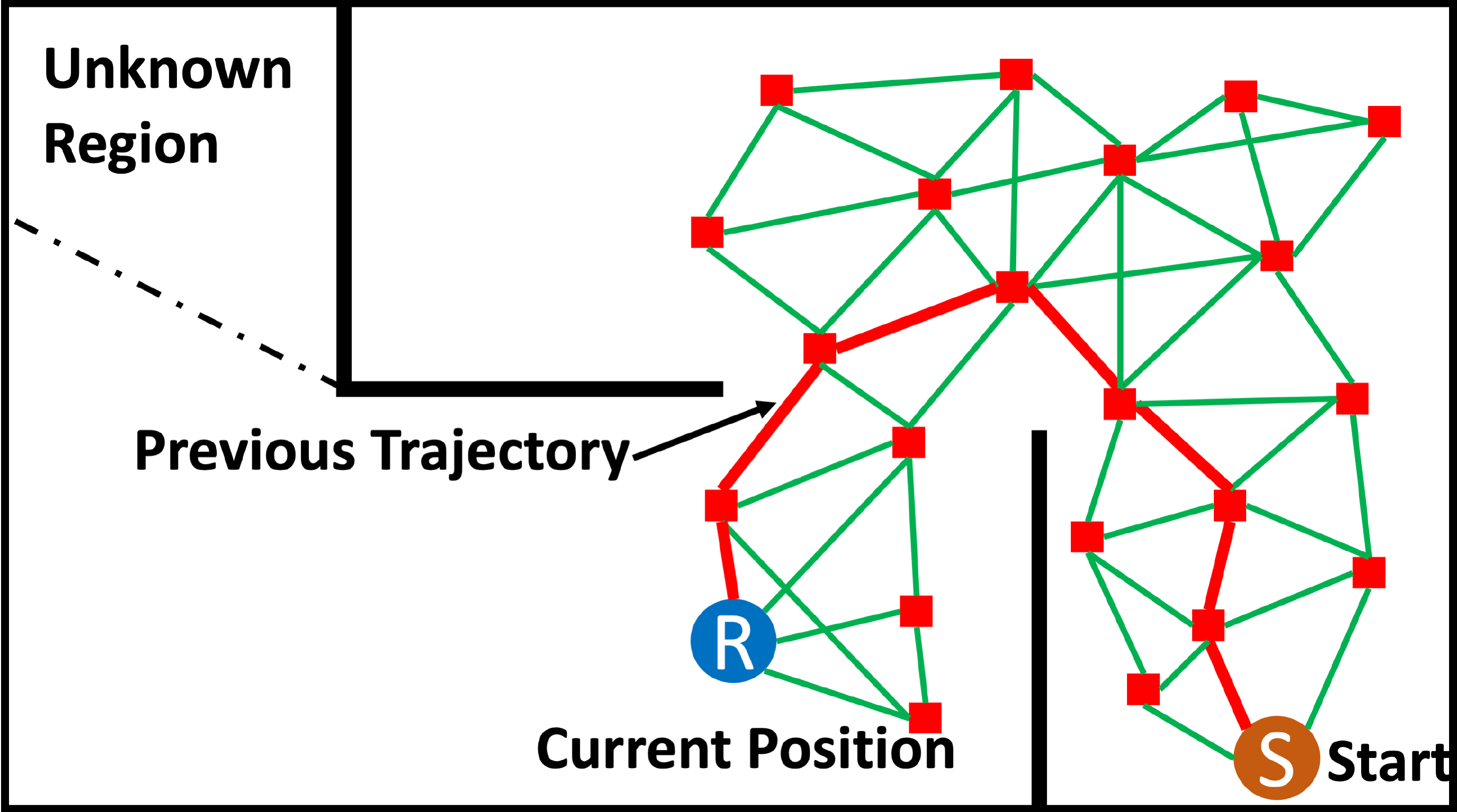}}\
\subfloat[Incremental new node sampling]{\includegraphics[width=.325\linewidth]{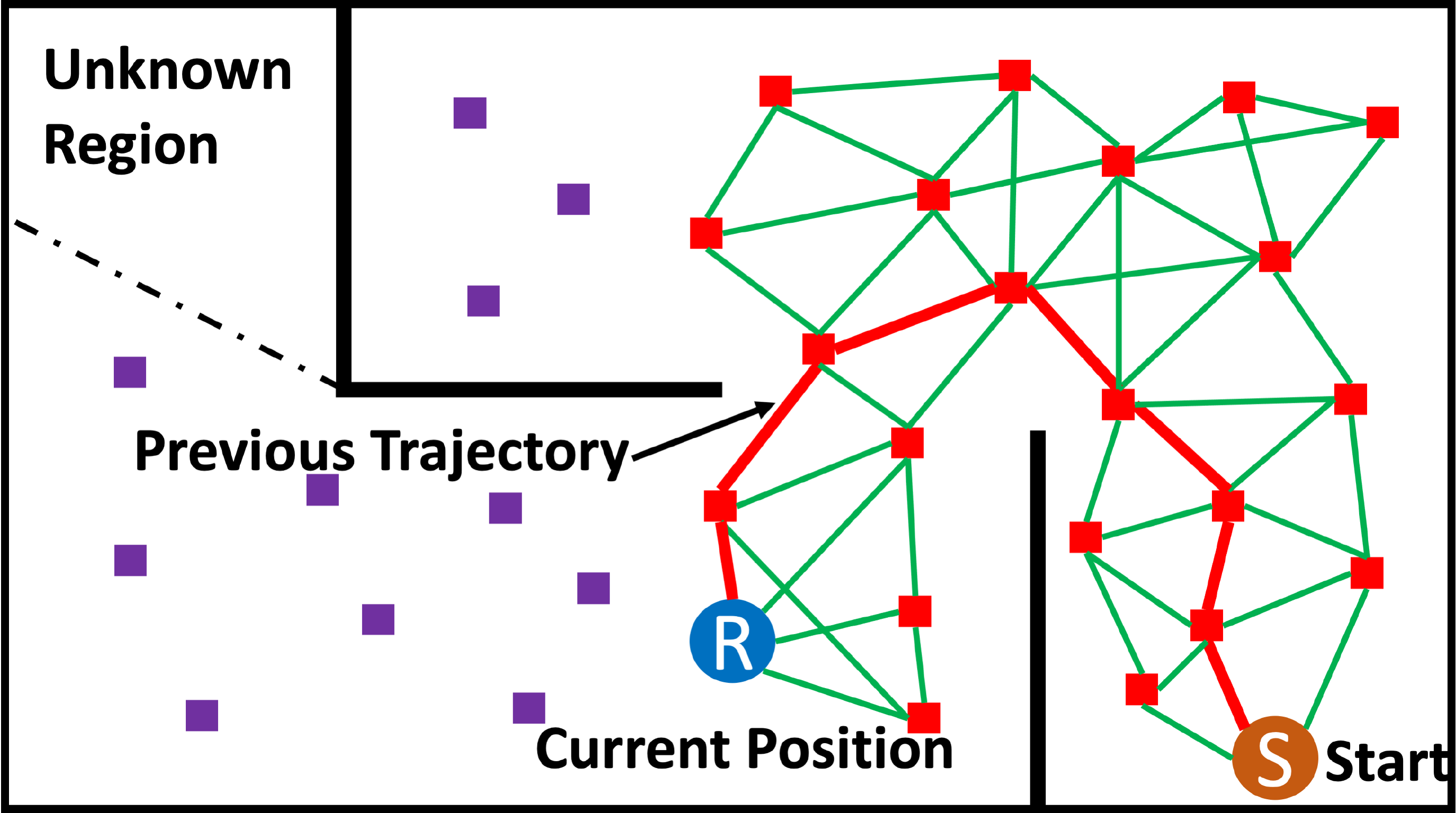}}\
\subfloat[New node connection]{\includegraphics[width=.32\linewidth]{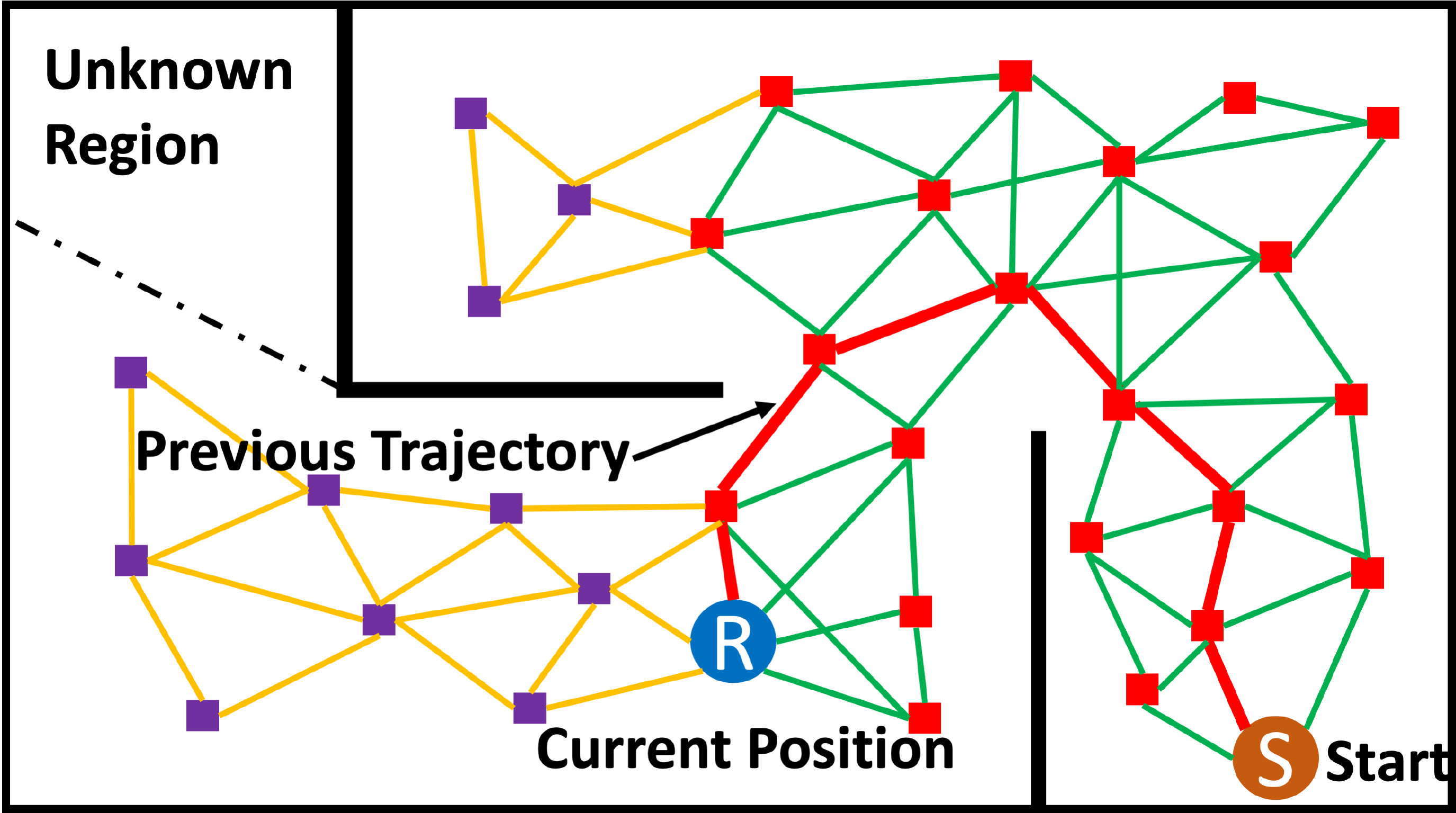}}
\caption{Illustration of the incremental roadmap construction. After completing the previous trajectory, new nodes (purple rectangles) are incrementally sampled, and new edges (yellow lines) are added to update the roadmap.}
\label{iPRM_illustration}
\end{figure*}

The planner can be divided into four steps shown in Fig. \ref{algorithm_flowchart} roadmap manager and trajectory module: (1) roadmap construction, (2) node evaluation and update, (3) trajectory generation, and (4) ESDF-based optimization. The first step incrementally adds nodes to the roadmap from the previous iterations while ensuring even distribution of nodes. This process is visualized in Fig. \ref{iPRM_illustration}. Then, the information gain of newly added nodes and the old nodes is evaluated or updated. After, the planner generates a trajectory based on the highest exploration scores and finally runs the optimization to further shorten the expected trajectory execution time and maintain the safety distance to obstacles. When encountering dynamic obstacles, the planner uses its current roadmap to generate the collision-free trajectory to shorten the replanning time.

\subsection{Roadmap Construction}
The incremental PRM needs to add nodes when a new region is observed. To better estimate the exploration utility of positions in the environment, roadmap nodes need to comprehensively cover the observed region. Besides, all the nodes need to be evenly distributed to ensure the roadmap quality. Simple random sampling will lead to higher density in previously observed regions and sparsity in newly observed regions. As such, we adopt a two-stage sampling strategy. The algorithm first takes random samples in the local region near the robot's current position and then samples in the global area. This sampling can ensure better coverage of the newly observed region around the robot. Instead of directly limiting the sampling number, our sampling terminates when the roadmap is globally \say{saturated.} Alg. \ref{sampling} shows our incremental PRM sampling strategy.  Whenever taking a random sample, we check the distance to its nearest neighbor (Line \ref{d_nn_l}). Only if the distance is greater than the threshold, then the node will be added into the roadmap; otherwise, we increase the sampling failure count $\mathcal{N}_\text{f}$ (Lines \ref{if_dist_l}-\ref{end_if_dist_l}). When the failure count is greater than predefined $\mathcal{N}_{\text{max}}$, the region is considered saturated.

Unlike the traditional PRM, our node connection is in a separate step, followed by the sampling (Alg. \ref{connection}). For each node from the previous sampling stage, we obtain nodes in its neighborhood. To connect two nodes, we consider three constraints: traversability, distance, and sensor range (Lines \ref{start_condition}-\ref{end_condition}). For the distance constraint, we want to avoid connecting two very far away nodes for roadmap quality. Moreover, to observe dynamic obstacles in the path segment, we need to make sure each connected node pair can observe each other using the robot sensor. 

\removelatexerror
\begin{algorithm}[t]
\caption{Roadmap Sampling - Global/Local}
\SetAlgoNoLine%
\label{sampling}
 $\mathcal{R} \gets$ \text{Roadmap}\;
 $\xi_\text{0} \gets$ \text{current robot pose}\;

 $\mathcal{S}_{\text{cond}} \gets false$ \Comment*[r]{saturation condition}
 \While{\normalfont{\textbf{not}} $\mathcal{S}_{\text{cond}}$}{
    $\mathcal{N}_{\text{fail}} \gets 0 $\;
    \While{true}{
        \If{$\mathcal{N}_{\text{fail}} > \mathcal{N}_{\text{max}}$}{
            $\mathcal{S}_{\text{cond}} \gets true$\;
            \textbf{break};\
        }
        $\mathcal{R}_{\text{s}} \gets \textbf{getSampleRegion}(\xi_\text{0})$\;
        $n_{\text{new}} \gets \textbf{randomConfig}(\mathcal{R}_{\text{s}})$\; 
        $d_{\text{nn}} \gets \textbf{disToNearestNeighbor}(n_{\text{new}})$\; \label{d_nn_l}
        \eIf{$ d_{\text{nn}} < d_{\text{th,min}}$}{ \label{if_dist_l}
            $\mathcal{N}_{\text{fail}} \gets \mathcal{N}_{\text{fail}} + 1$;\label{end_if_dist_l} \
        }{
            Add $n_{\text{new}}$ to $\mathcal{R}$;\  
        }
    }
 }
\end{algorithm}

\begin{algorithm}[t]
\caption{Node Connection}
\label{connection}
\SetAlgoNoLine%
	\For {Node $n_{\text{i}}$ \normalfont{\textbf{in}} Roadmap $\mathcal{R}$}{
	    $N \gets \textbf{neighborHood}(n_{\text{i}})$;\\
	    \For {$n_{\text{N}}$ \normalfont{\textbf{in}} N}{
	        $\mathcal{C}_{\text{collision}} \gets \textbf{checkCollision}(n_{\text{i}}, n_{\text{N}})$;\\ \label{start_condition}
	        $\mathcal{C}_{\text{dis}} \gets \textbf{distanceTo}(n_{\text{i}}. n_{\text{N}}) \leq d_{\text{th,max}}$;\\
	        $\mathcal{C}_{\text{sensor}} \gets \textbf{rangeCondition}(n_{\text{i}}. n_{\text{N}})$;\\ \label{end_condition}
	        \If {$\mathcal{C}_{\text{collision}} \normalfont{\textbf { and }} \mathcal{C}_{\text{dis}} \normalfont{\textbf{  and  }} \mathcal{C}_{\text{sensor}}$}{
	            $\textbf{connect}(n_{\text{i}}, n_{\text{N}})$;\
	        }
	    }
	 }
\end{algorithm}

\subsection{Node Evaluation and Update Rule}
After incrementally constructing the roadmap, node evaluation and update take place. Similar to the information gain proposed in \cite{5}, our node gain is based on the number of expected unknown visible voxels within the sensor's FoV. Also, instead of only computing the number of voxels of a specific yaw angle, we discretize the orientation angle into $N$ parts and, for each part, calculating the corresponding number of voxels. Then, we use the total number of voxels to represent the node gain. This node gain represents the general exploration utility, and the gain of each orientation is stored for later interpolation in the trajectory generation. Besides, we divide the unknown voxels into three categories: normal unknown, frontier unknown, and surface unknown. The surface unknown must be adjacent to both free and occupied voxels, while the frontier unknown only needs to have the adjacent free voxel. We assign the highest weight for the surface unknown voxels as they are more valuable in inspecting the surface and observing the environment's contour. The frontier unknown voxels have the second-highest weight since they are more reliable than normal unknown voxels. The formula for computing the node gain is defined as:

\begin{equation}
\label{gain}
\textbf{Gain}(n) = w_\text{n} \cdot \mathcal{U}_{\text{n,tot}} + w_\text{f} \cdot \mathcal{U}_{\text{f,tot}} + w_\text{s} \cdot \mathcal{U}_{\text{s,tot}},
\end{equation}

\noindent where $w_\text{n}, w_\text{f},$ and $w_\text{s}$ denotes the weights for normal unknown, frontier unknown, and surface unknown voxels, respectively. $\mathcal{U}_{\text{n,tot}}, \mathcal{U}_{\text{s,tot}},$ and $\mathcal{U}_{\text{f,tot}}$ represent the estimated number of unknown voxels within the sensor's range for each type of unknown voxel. The node gain is the weighted sum of the number of unknown voxels for normal, frontier, and surface unknown voxels, respectively. 

Since it is inefficient to re-compute the gain for all the previous nodes in each iteration, we apply a rule to determine which subset of nodes should be updated. We first record a set of nodes, $S_{n}$, near the previous robot's trajectory, which can be defined in Euclidean distance. In the second step, from the recorded set, $S_{n}$, we selected the nodes whose gain value or distance to the previous trajectory is lower than the threshold and set their gain value to zero without reevaluation. From our experiment observation, these nodes often have a low exploration utility and can be neglected for exploration goal selection. Finally, for the rest of the nodes, we apply the formula in Eq. (\ref{gain}) to recalculate their gain value. Usually, at this step, there are only 10-20\% of nodes left in the recorded set, $S_{n}$. During the exploration process, the known region nodes tend to have zero information gain. In contrast, nodes in the edge and border between known and unknown usually have the highest gain.

\subsection{Trajectory Generation}
To minimize the total exploration time, we generate a trajectory with the highest information gain rate. The information gain rate is the expected gain per unit time, and we use it as the score to evaluate trajectories. Based on the node gain, we collect a set of goal candidates, $G_{\text{c}}$. The ratio of the minimum gain value to the maximum value of the goal in the set cannot be less than the value, $0<\lambda<1$:
\begin{equation}
\label{goal_candidates}
G_{\text{c}} = \{n_{\text{i}} \in \textbf{Node}(\mathcal{R}) \|\ \textbf{Gain}(n_{\text{i}}) \geq \lambda \cdot \textbf{Gain}(n_{\text{max}})\} ,\\
\end{equation}
\noindent where $\mathcal{R}$ is the roadmap, and $n_{\text{max}}$ is the node with highest gain value. For every node in the roadmap, we only collect the nodes that have the gain value greater than the threshold, $\lambda \cdot \textbf{Gain}(n_{\text{max}})$. Then, graph search algorithms such as A* and Dijkstra are used for finding the shortest path. Each trajectory score is evaluated by the summation of the expected gain from each node divided by the total execution time:

\begin{equation}
\text{Trajectory Score} = \frac{\sum_{i=1} \textbf{Gain}(n_{\text{i,yaw}})}{\text{Execution Time}}. \\
\end{equation}

\noindent Note that the yaw angle is automatically determined when we obtain the waypoint nodes for trajectory dynamic obstacle detection. We use each orientation angle's previously saved gain to interpolate the expected gain value for a specific yaw angle. The trajectory with the highest score is selected for execution. The execution time in the denominator is calculated based on the predefined velocity and acceleration of the robot.

The generated trajectory ensures the robot to detect the unexpected dynamic obstacle in its moving path segment. When a potential collision is predicted, the new trajectory is generated accordingly with a higher $\lambda$ value to speed up the replanning. Since no 
nodes are added, and no information gain needs to be recalculated, the robot can immediately obtain a new alternative trajectory to avoid dynamic obstacles safely.

\subsection{ESDF-based Optimization}
To further shorten the exploration time and path length, and to increase the safety distance to obstacles, we formulate the ESDF-based optimization. Our objective includes both the trajectory execution time and the trajectory's average distance to obstacles. From the ESDF map, we can obtain the minimum distance to obstacles for each node. Since the trajectory is already generated with high information gain, we only run the local region's optimization for each node. Formally, the optimization problem is formulated as follow:\\
Minimize:
\begin{equation}
\label{objective_function}
 F(\textbf{n}) = w_\text{t} \cdot \frac{t(\textbf{\text{n}})}{t_\text{0}} + w_\text{d} \cdot \frac{d_\text{0}}{d(\textbf{\text{n}})}, \ \textbf{n} = \{n_\text{1},.. ,n_\text{N}\},
\end{equation}
subject to:
\begin{subequations}
\begin{align}
\prod_{i=1}^{N-1} C_{n_{\text{i}}, n_{\text{i+1}}} \cdot S_{n_{\text{i}}, n_{\text{i+1}}} \cdot D_{n_{\text{i}}, n_{\text{i+1}}} = 1 \label{constraint1}, \\
n_i = \textbf{Local}(n_{\text{i, 0}}), \ \forall n_\text{i} \in \textbf{n} \label{constraint2}, \\
C_{n_{\text{i}},{n_{\text{i+1}}}}, \ S_{n_{\text{i}},{n_{\text{i+1}}}}, \ D_{n_{\text{i}},{n_{\text{i+1}}}} \in \{0, 1\},
\label{constraint3}
\end{align}
\end{subequations}

\noindent where \textbf{n} represents the waypoints in the trajectory. Note that Eq. \ref{objective_function} normalizes both trajectory time and obstacle distance by the initial value from the non-optimized trajectory. Weight factors, $w_\text{t}$ and $w_\text{d}$, are applied to balance the importance of the different objectives. In Eq. \ref{constraint1} and \ref{constraint3}, $C_{n_{\text{i}}, n_{\text{i+1}}}, S_{n_{\text{i}}, n_{\text{i+1}}},$ and $D_{n_{\text{i}}, n_{\text{i+1}}}$ represent collision, sensor range, and minimum distance condition for two connected nodes, and each condition has a binary value (Eq. \ref{constraint3}), where 1 indicates the condition being satisfied and 0 otherwise. Eq. \ref{constraint1} considers the trajectory collision, sensor range, and minimum distance. As mentioned before, the next node should be selected within the sensor range of its previous node for obstacle detection. Also, we expect two nodes are not too close to each other so that nodes do not overlap their corresponding gain voxels. The local optimization is reflected in Eq. \ref{constraint2}, where each node has to be in the local region of the initial value.

\begin{figure}[t]
    \centering
    \includegraphics[scale=0.32]{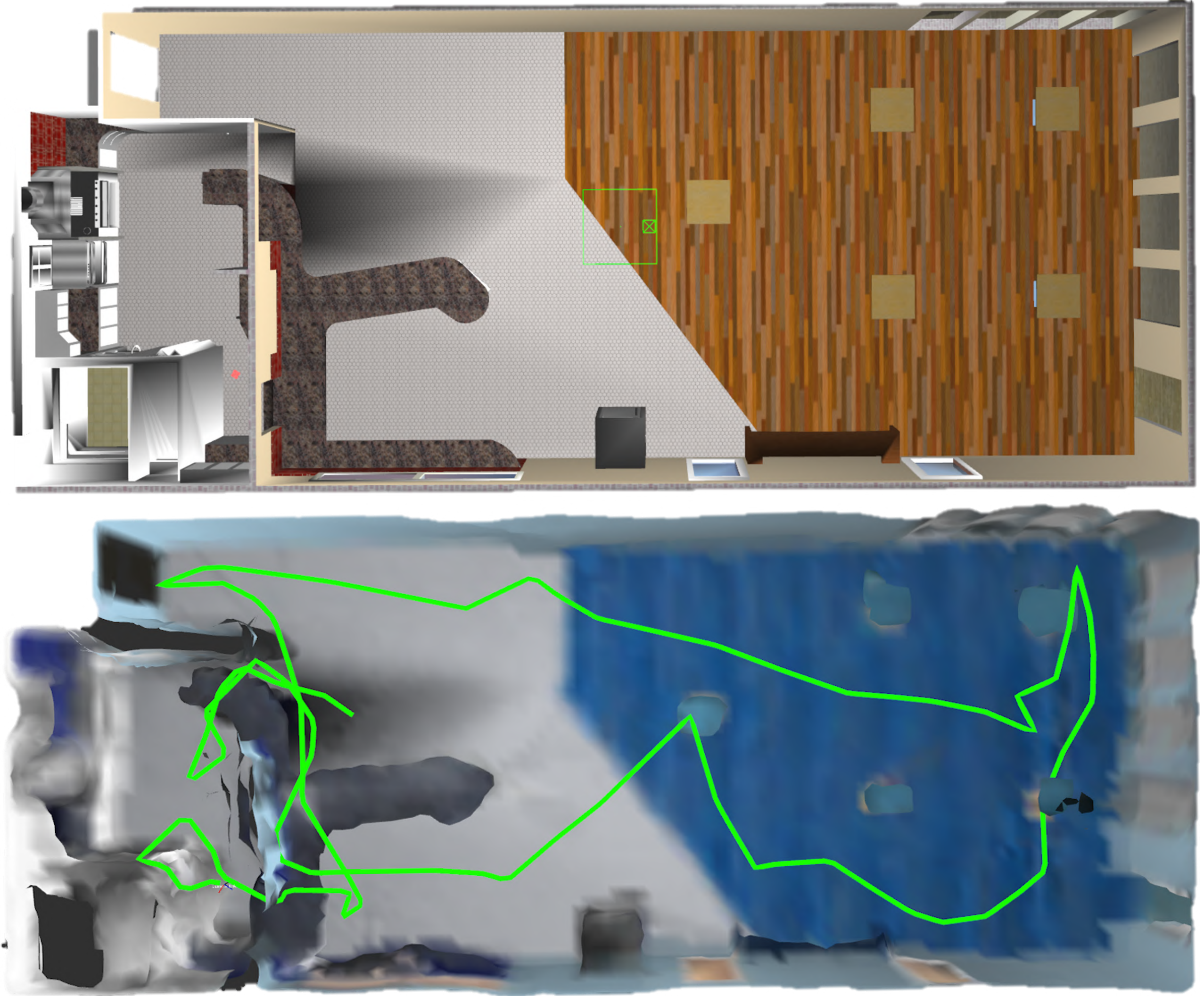}
    \caption{Visualization of the Cafe (Small) environment with its fully explored map and exploration trajectory.}
    \label{fig:cafe_gazebo}
\end{figure}

\begin{figure*}[t]
\centering
\subfloat[Frontier Exploration \cite{2}]{\includegraphics[width=.242\linewidth]{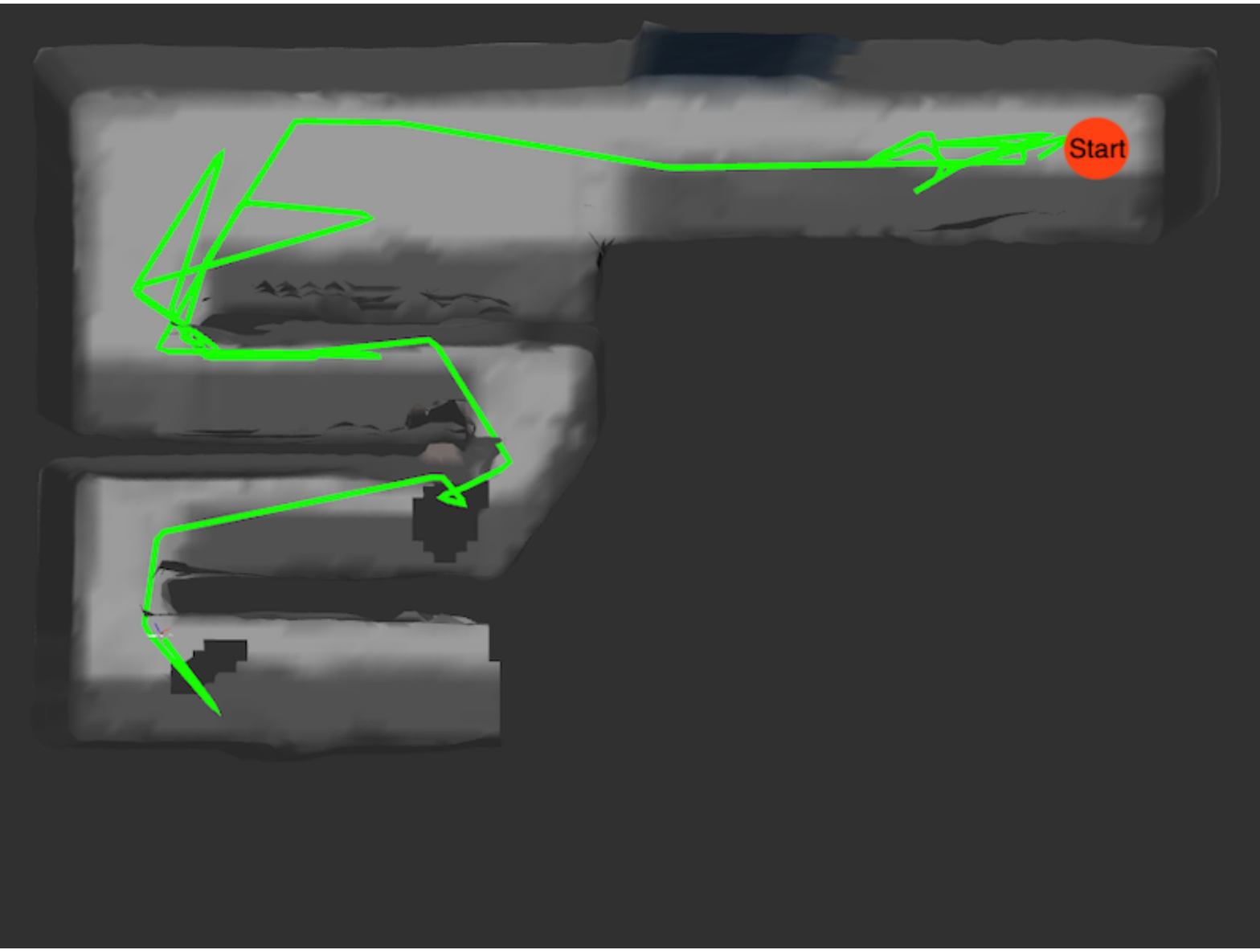}}\
\subfloat[RH-NBV \cite{5}]{\includegraphics[width=.242\linewidth]{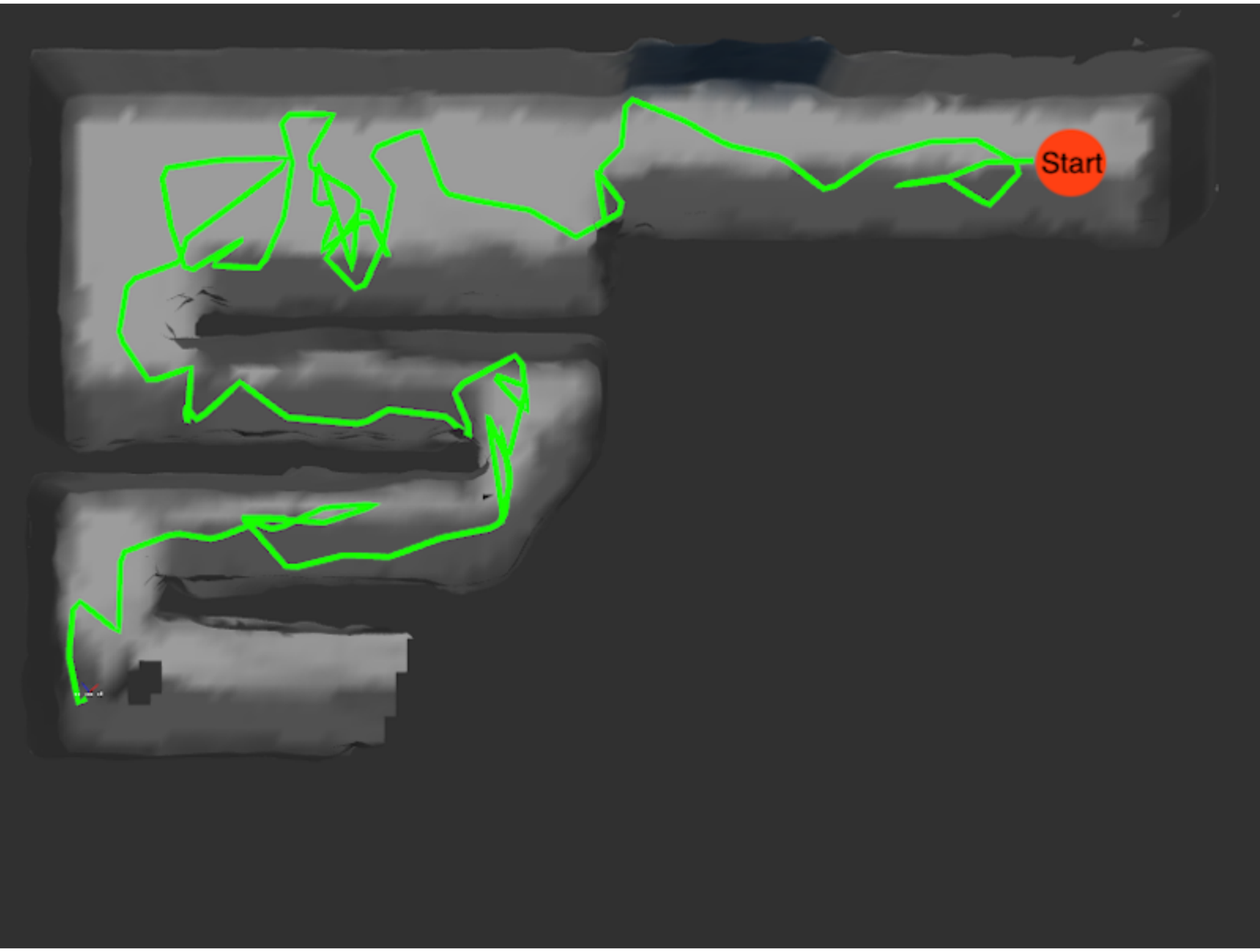}}\
\subfloat[AEP \cite{7}]{\includegraphics[width=.242\linewidth]{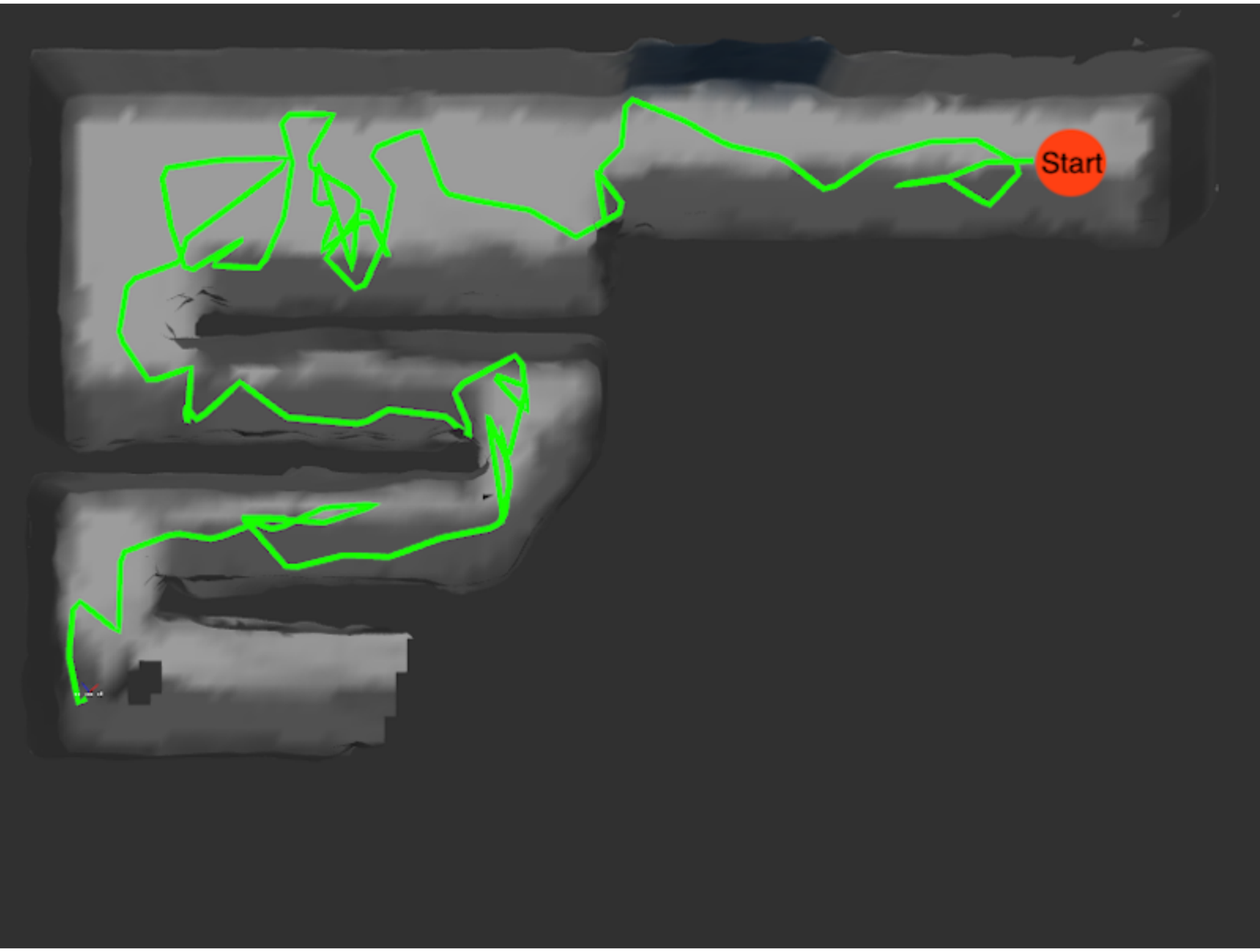}}\
\subfloat[DEP (Ours)]{\includegraphics[width=.242\linewidth]{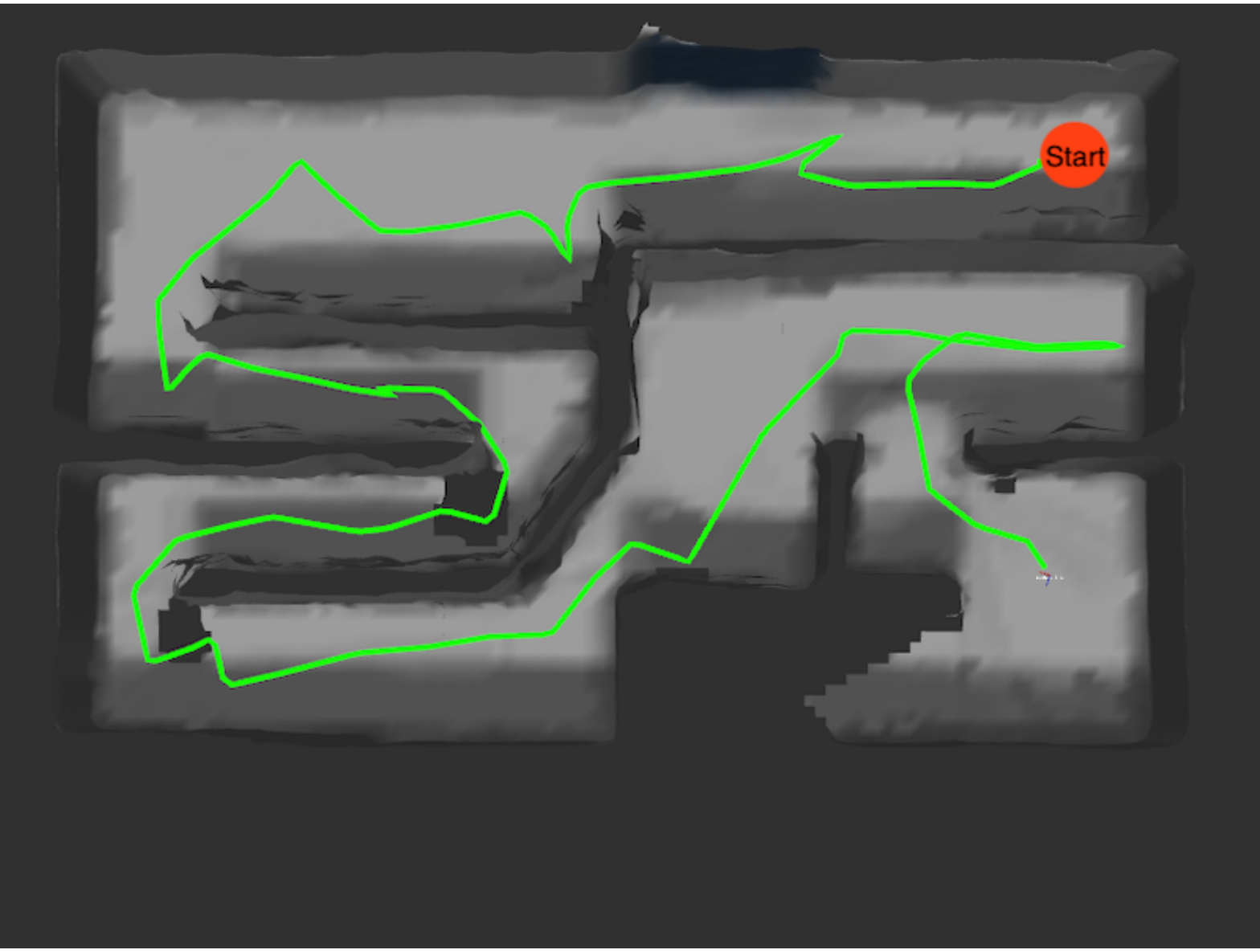}}
\caption{The explored map in the Maze (Medium) environment after running each planner for 10 minutes. The red dots and green lines are the robot's start positions and trajectories. Our planner has the largest explored area.}
\label{maze_screenshot}
\end{figure*}

\begin{figure*}[t]
\centering
\subfloat[Cafe (Small) Env. Exploration Rate]{\includegraphics[width=.325\linewidth]{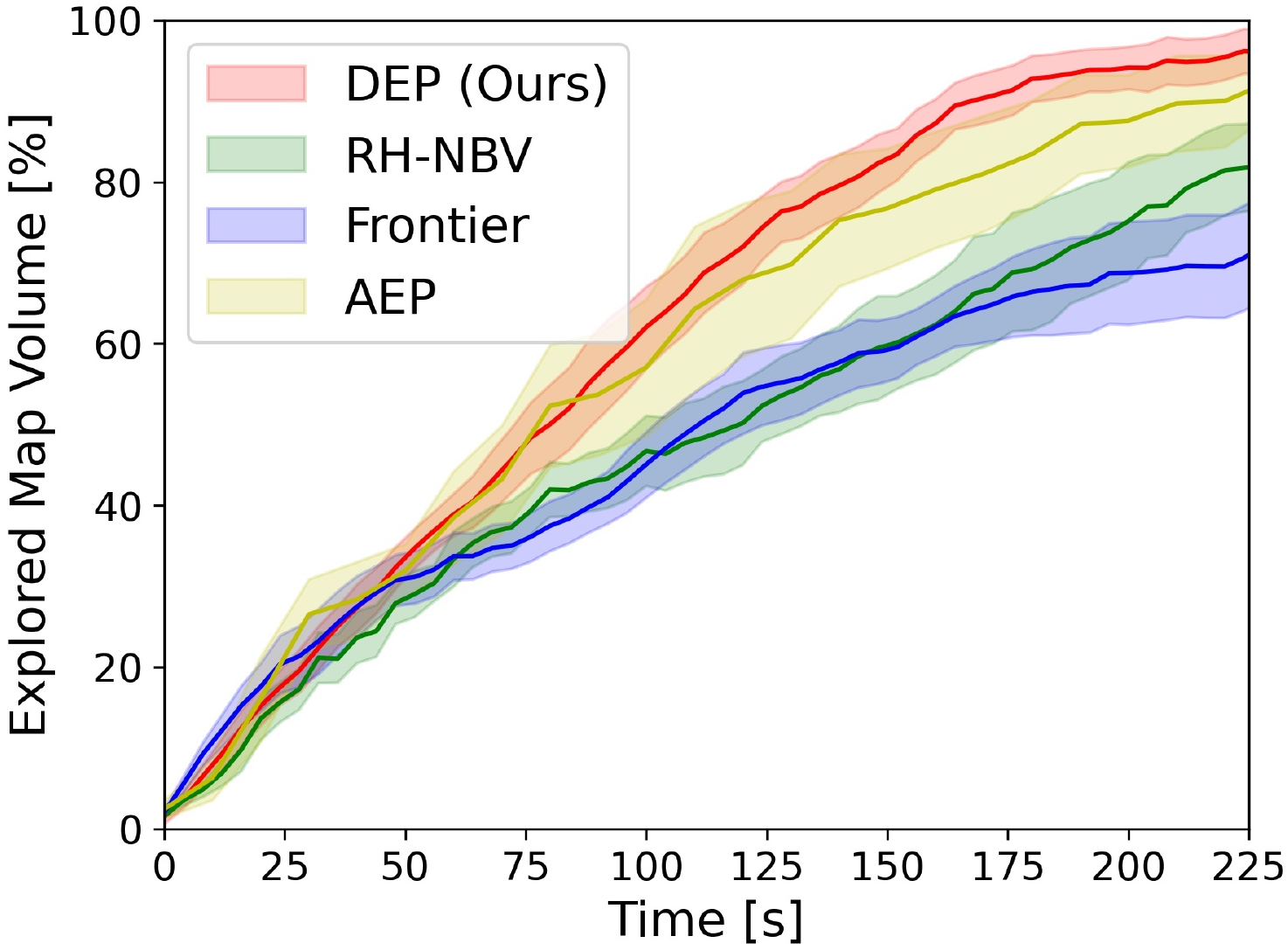}}\
\subfloat[Maze (Medium) Env. Exploration Rate]{\includegraphics[width=.325\linewidth]{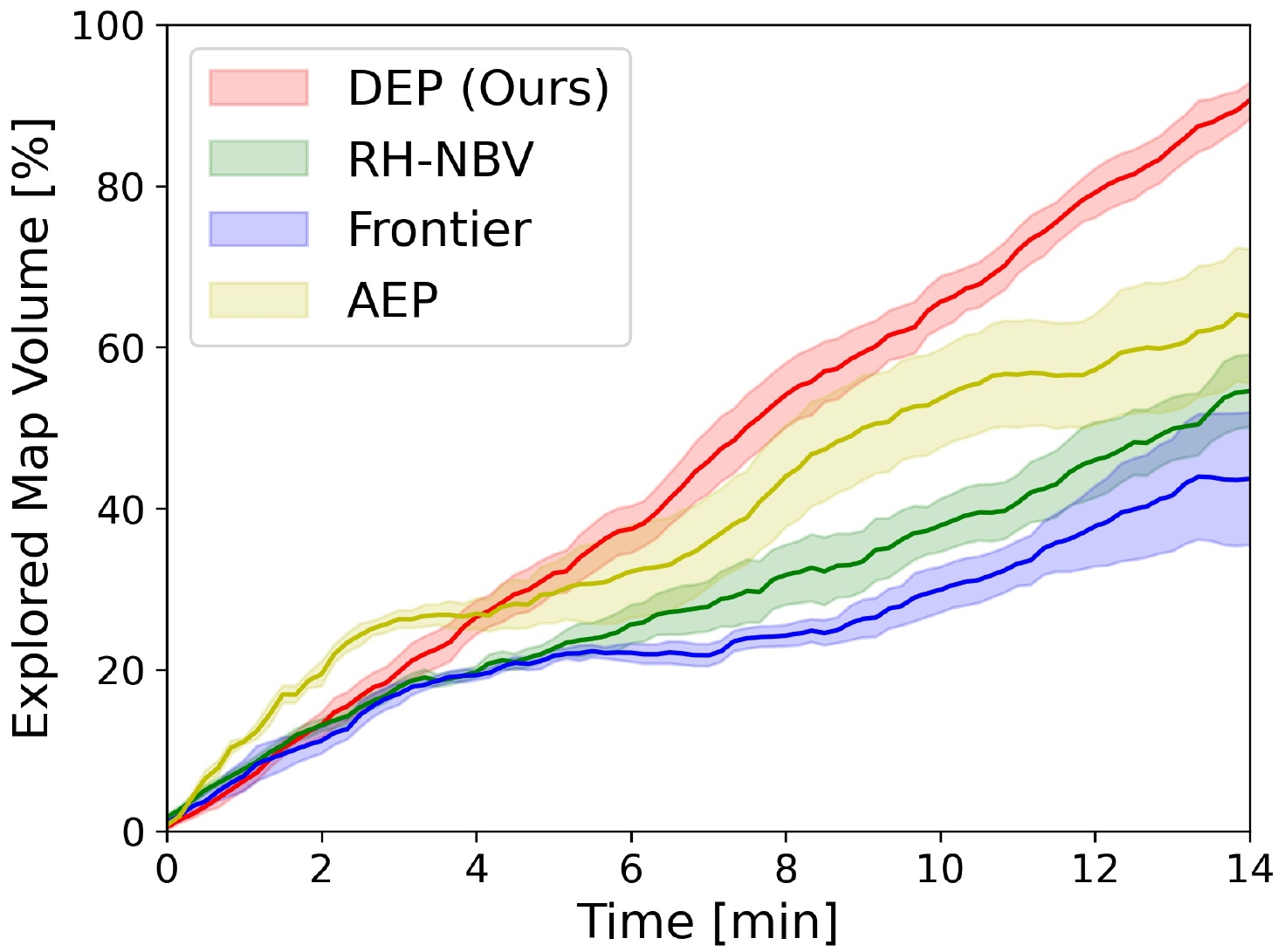}}\
\subfloat[Office (Large) Env. Exploration Rate]{\includegraphics[width=.325\linewidth]{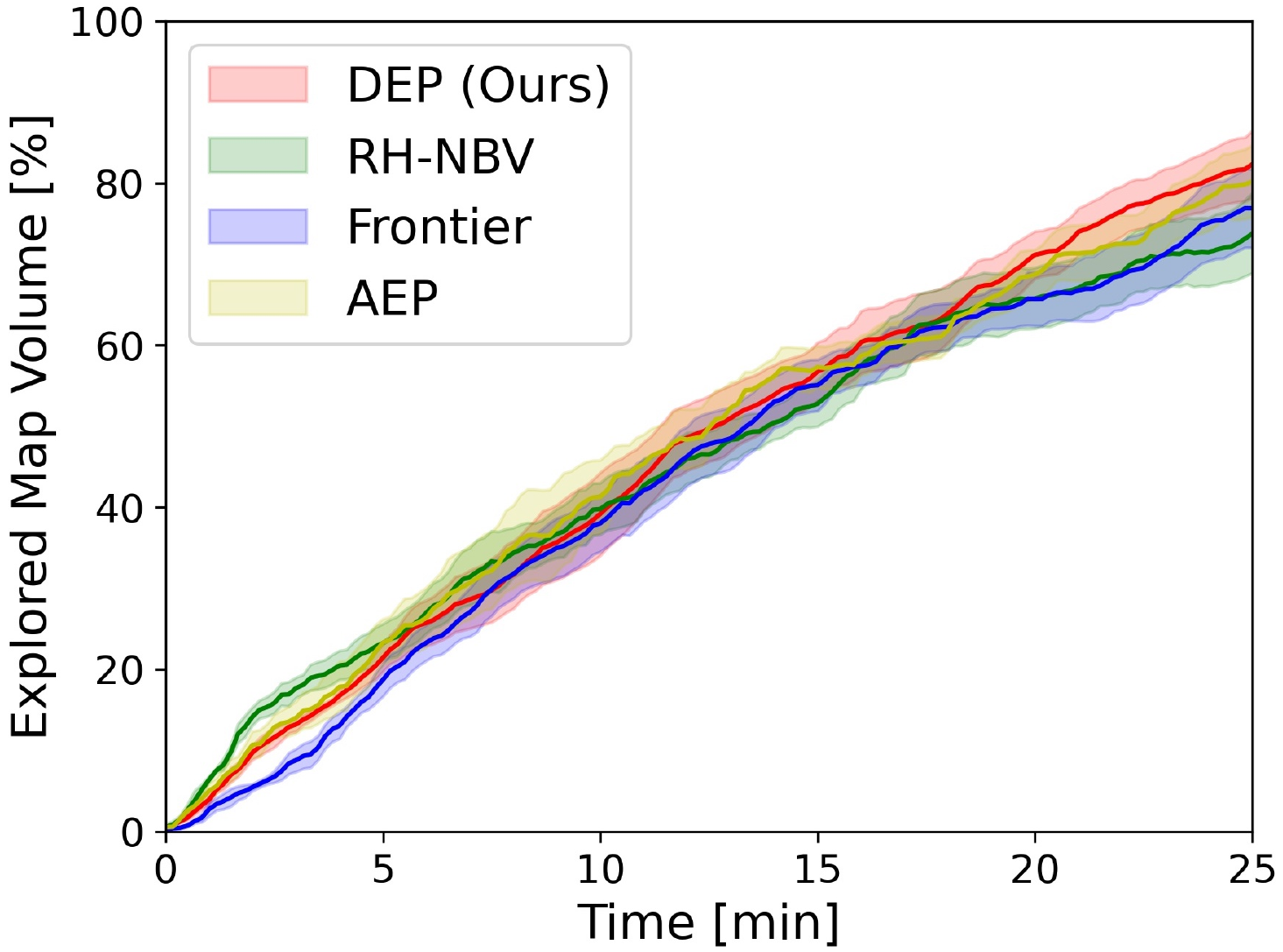}}
\caption{Comparison of the exploration rate in the Cafe (Small), Maze (Medium), Office (Large). Our planner has the highest exploration rate in the Cafe and Maze environments. Four planners have similar exploration rates in the Office.}
\label{comparison_figure}
\end{figure*}

\section{Implementation Details}
We use the Octomap \cite{23} package for planner implementation.  As \cite{5} suggests, the sensor range used for the planner to calculate the information gain is less than the actual maximum sensor range, $d_{\text{planner}} < d_{\text{max}}$. The goal node's yaw angle optimization is similar to \cite{6}. For the rest of the nodes, the yaw must be the robot's moving direction to maximize the visible range for dynamic obstacle detection. The termination criterion is defined as no new nodes added for the consecutive $N$ steps, which implies no unknown regions left for exploration. For the ESDF-based optimization, we use the Voxblox \cite{24} package to generate the TSDF/ESDF map. The local region of each node is encoded by a predefined bounding box with a given side length. The average distance to the obstacle is computed by discretizing the trajectory with the same resolution as the TSDF/ESDF map. We terminate the optimization process after $N_{opt}$ iterations to ensure fast planning.

\section{Results and Performance Benchmarking}
To fully analyze the presented algorithm's performance, we conduct the simulation experiments using a quadcopter equipped with an RGB-D camera. The system is running on Intel i7-7700HQ at 2.4 GHz. The experiments are to evaluate the exploration performance and safety. For exploration performance,  we choose RH-NBV \cite{5}, the frontier-based exploration algorithm mentioned in \cite{25} and the most recent AEP \cite{7} as our benchmarks. The video of all the experiments can be found on: \url{https://youtu.be/ileyP4DRBjU}

\begin{table}[t]
\renewcommand\arraystretch{1.2}
\begin{center}
\caption{Parameters for our DEP planner.} \label{planner_parameters}

\begin{tabular}{ |l|c| } 
\hline

Parameter & Value \\[0.5ex]
\hline

Node Min. Distance,\ $d_{\text{th,min}}$ & 0.8 m \\ 

Node Max. Distance,\ $d_{\text{th,max}}$  & 1.5 m \\ 

Max. Sample Failure,\ $\mathcal{N}_{\text{max}}$ & 50  \\ 

Gain Parameters $[w_\text{n}, w_\text{f}, w_\text{s}]$ & $[1, 2, 4]$ \\

Gain Cutoff Thresh. (Regular),\ $\lambda_\text{0}$ & 0.5 \\

Gain Cutoff Thresh. (Replan),\ $\lambda_\text{r}$ & 0.8 \\

Optimization Range Box & [0.5, 0.5, 0.5]$\text{m}^\text{3}$ \\

\hline
\end{tabular}
\end{center}
\end{table}

\subsection{Environments and Parameters}
We select a total of six target environments for experiments, as shown in TABLE \ref{env_info}. We use the environments with different sizes, Cafe (Small), Maze (Medium), and Office (Large), to test the overall exploration performance. To demonstrate the ability to generate safe paths with dynamic obstacles, we include three dynamic environments: Dynamic Auditorium, Dynamic Tunnel, and Dynamic Field, which contains several walking people whose trajectories are unknown to the robot. 

The planner and robot parameters are presented in TABLE \ref{planner_parameters} and \ref{experiment_parameters}, respectively. Those parameters are selected to achieve: (i) the evenly distributed roadmap, (ii) the efficient and reliable gain estimation, (iii) the fast replanning speed, and (iv) reasonable optimization constraints. For benchmarks, we apply the suggested parameters from their works \cite{5} \cite{25} \cite{7}. For fairness, the robot parameters are the same for each planner.

\begin{figure}[t]
    \centering
    \includegraphics[scale=0.3]{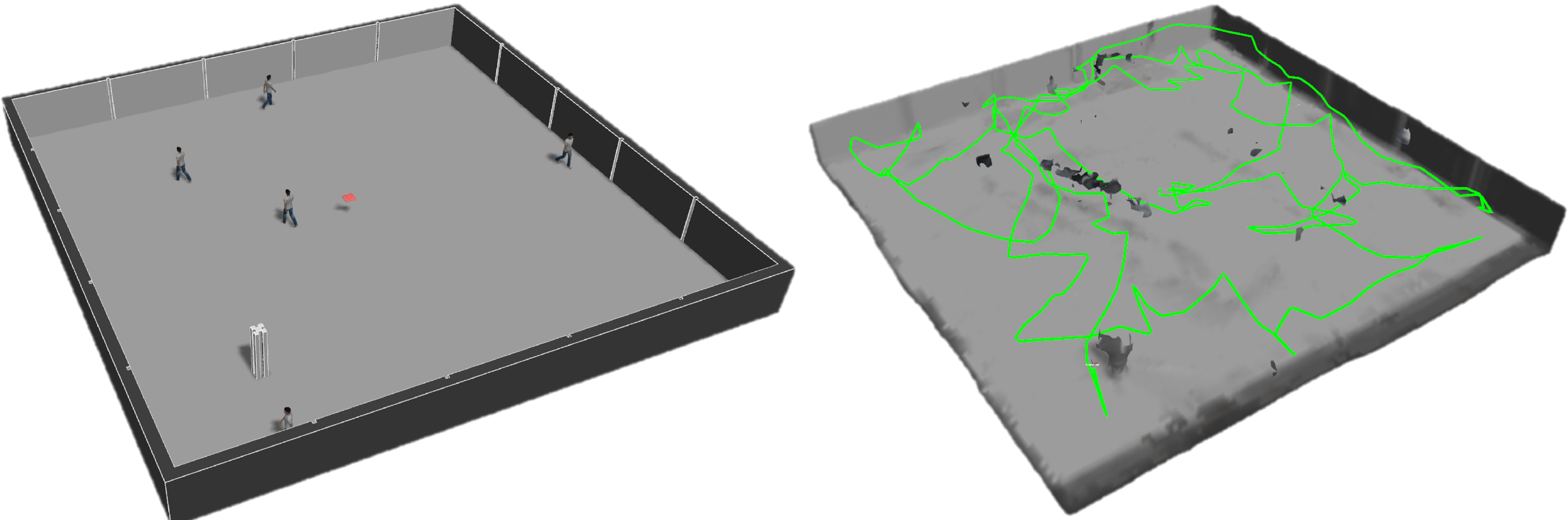}
    \caption{Visualization of the Dynamic Field environment with its fully explored map and exploration trajectory.}
    \label{fig:dynamic field}
\end{figure}

\begin{table}[t]
\renewcommand\arraystretch{1.2} 

\begin{center}
\caption{Information of the target environments.} \label{env_info}
\begin{tabular}{ |c|c|c| } 
 \hline

 Env No. & Environment Name & Size ($\text{m}^\text{3}$) \\
 \hline

 1 & Cafe (Small) & $20 \times 10 \times 3$ \\
 \hline

 2 & Maze (Medium) & $20 \times 20 \times 3$ \\ 
 \hline

 3 & Office (Large) & $40 \times 30 \times 3$ \\  
 \hline

 4 & Dynamic Auditorium & $20 \times 15 \times 4$ \\  
 \hline

 5 & Dynamic Tunnel & $20 \times 25 \times 9$ \\ 
 \hline

 6 & Dynamic Field & $24 \times 24 \times 3$ \\
 \hline
\end{tabular}
\end{center}
\end{table}

\begin{table}[t]
\renewcommand\arraystretch{1.2}
\begin{center}
\caption{Robot (Quadcopter) parameters and settings.} \label{experiment_parameters}

\begin{tabular}{l  l  l  l} 
\hline

Max. Linear Vel. & 0.3 m/s & Collision Box & [0.4, 0.4, 0.3]$\text{m}^\text{3}$\\

Max Angular Vel.  & 0.8 rad/s & Camera Range & 4 m\\ 

Map Resolution & 0.2 m & Camera FOV & [103.2, 77.4]\degree \\  

\hline
\end{tabular}
\end{center}
\end{table}

\subsection{Exploration Performance}
We record the total exploration time, path length, and computational time for each planner and take ten experiments to obtain the means and standard deviations. 


TABLE \ref{exp_comparison} presents the overall exploration performance comparison in three different-scale environments. In the Cafe (Small), our planner (DEP) has the least exploration time (4.77min), total path length (43.12m), and computational time (0.17min). AEP \cite{7}, as the improvement of the frontier exploration and RH-NBV, outperforms those two planners in exploration time and has similar exploration path length and computational time. However, compared to DEP, AEP still spent 14.9\% and 117.6\% longer time to finish exploration and perform computation, respectively, and also takes 37.2\% longer path. In the Maze (Medium), DEP still has the best performance, and the second-best planner AEP takes 23.23 minutes for exploration, which is 30.9\% longer than DEP's 17.75 minutes. Frontier exploration and RH-NBV take 34.74 and 31.44 minutes for exploration, respectively, much longer than DEP's. In the Office (Large), DEP has the shortest exploration and computational time, while frontier exploration has the least total path length, 253.63m, which is 20.3\% shorter than our DEP's 318.53m. It is worth noting that, in Office (Large) environment, DEP's computational time is significantly shorter: 36.9 \% of the time as AEP and nearly 20\% of the time as frontier exploration and RH-NBV.   

The exploration rate is shown in Fig. \ref{comparison_figure}. In the Cafe (Small), our planner (DEP) has the highest exploration rate after 50 seconds, slightly higher than AEP's. RH-NBV and frontier exploration have the lowest rates throughout the process. In the Maze (Medium), the gap between each planner is larger. DEP almost keeps the highest rate during the exploration as its slope does not decrease, while the other three planners have different extent rate drops after a particular time. AEP's rate outperforms frontier exploration and RH-NBV during the exploration. To visualize the exploration process, we run four planners for 10 minutes under the same settings and record their corresponding maps and trajectories, as shown in Fig. \ref{maze_screenshot}. Our DEP has the largest explored area among four planners. AEP is better than frontier exploration and RH-NBV since frontier exploration has many back-and-forth trajectories, and RH-NBV's trajectory is crooked. However, in the Office (Large) environment, three planners have a similar exploration rate at all stages except the very end, when our DEP's exploration rate is slightly higher than the other three's.  

\subsection{Exploration in Dynamic Environments}
\begin{figure}[t]
    \centering
    \includegraphics[scale=0.3]{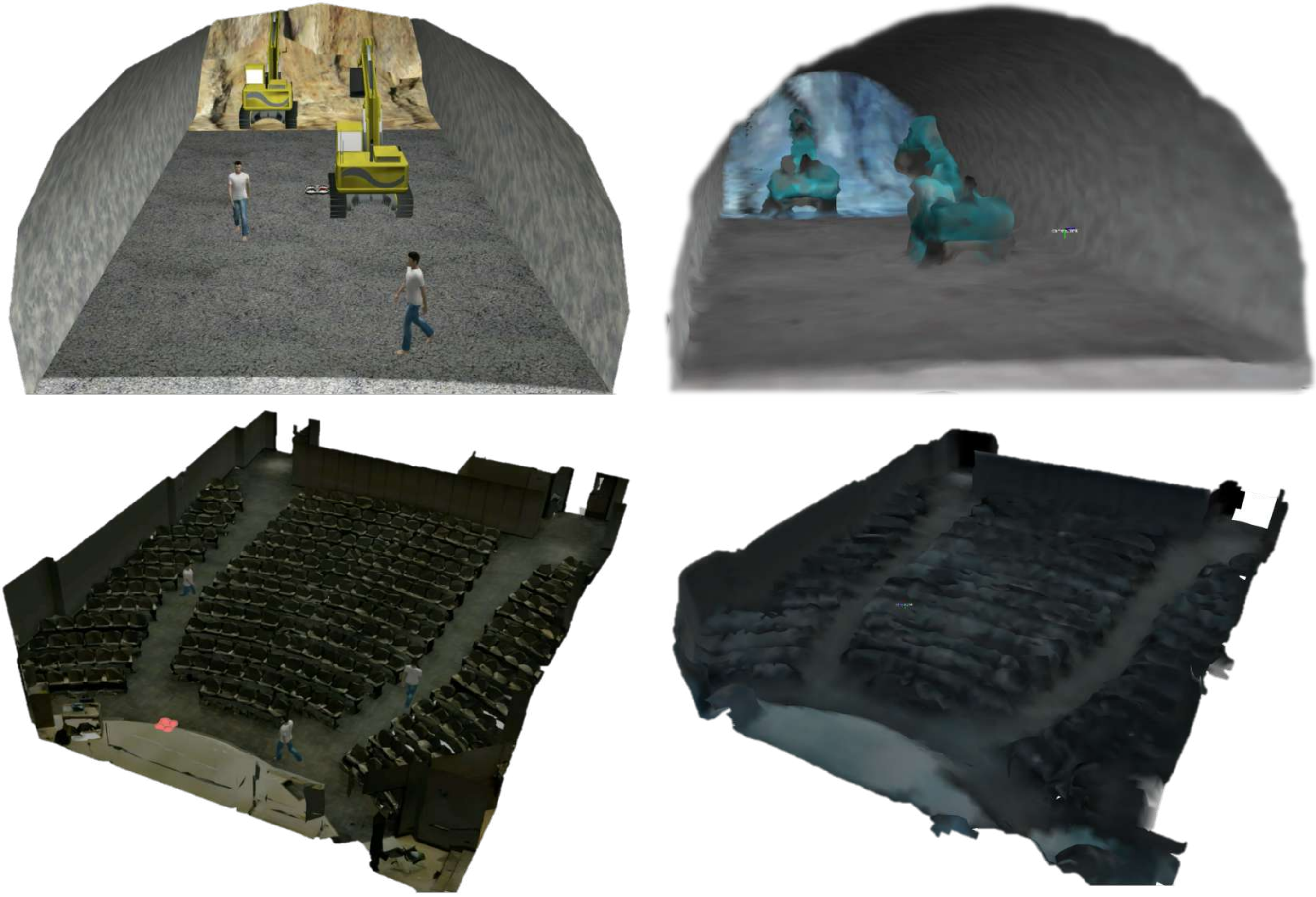}
    \caption{Visualization of dynamic tunnel and dynamic auditorium environments (left) with their fully explored maps (right)}
    \label{fig:tunnel and auditorium}
\end{figure}

To prove the ability of safe exploration with dynamic obstacles, Dynamic Auditorium, Dynamic Tunnel, and Dynamic Field are used to test our planner. Each environment includes several walking people whose trajectories are unknown to the robot. We run ten experiments and check whether collision happens or not. The results show that, in all three dynamic environments, our planner has no collision with dynamic obstacles and can safely and fully explore each environment. Fig. \ref{fig:dynamic field} and Fig. \ref{fig:tunnel and auditorium} give the visualization of each environment with the fully explored map.

\begin{table}[t]
\renewcommand\arraystretch{1.2} 
\begin{center}
\caption{The ratio of the expected trajectory execution time, path length, and distance to the nearest obstacle for optimized trajectory and non-optimized trajectory at each planning iteration for different environments.} \label{optimization_comparison_table}

\begin{tabular}{|>{\centering\arraybackslash}m{1.3cm} || >{\centering\arraybackslash}m{1.5cm} >{\centering\arraybackslash}m{1.5cm} >{\centering\arraybackslash}m{2cm}|} 
\hline
 Env. Name & Time Ratio $\frac{T_{\text{opt}}}{T_{\text{non}}}$ & Length Ratio $\frac{L_{\text{opt}}}{L_{\text{non}}}$ & Distance Ratio $\frac{D_{\text{opt}}}{D_{\text{non}}}$\\ 
 
 \hline
 \hline
 Cafe & 87.44\% & 92.58\% & 112.95\% \\
 Maze & 88.43\% & 92.78\% & 127.46\% \\
 Office & 89.41\% & 92.90\% & 125.71\% \\
 Auditorium & 86.86\% & 91.49\% & 116.88\% \\
 Tunnel & 84.25\% & 89.55\% & 119.88\% \\
 \hline
 \textbf{Average} & \textbf{87.28\%} & \textbf{91.86\%} & \textbf{120.58\%} \\
\hline
\end{tabular}
\end{center}
\end{table}

\begin{table*}[t]
\renewcommand\arraystretch{1.2}
\begin{center}
\caption{Comparison of the exploration performance in the Cafe (Small), Maze (Medium), Office (Large) Environments.}

\begin{tabular}{ |>{\centering\arraybackslash}m{1.5cm} >{\centering\arraybackslash}m{1.5cm}||>{\centering\arraybackslash}m{1.5cm} >{\centering\arraybackslash}m{1.5cm}||>{\centering\arraybackslash}m{1.5cm} >{\centering\arraybackslash}m{1.5cm}||>{\centering\arraybackslash}m{1.5cm} >{\centering\arraybackslash}m{1.5cm}|}

 \hline

 \multicolumn{8}{|c|}{Exploration Performance in Different Environments} \\
 \hline

 \multicolumn{2}{|c||}{}  & \multicolumn{2}{c||}{Exploration Time (Min.)} &\multicolumn{2}{c||}{Total Path Length (m)}& \multicolumn{2}{c|}{Computational Time (Min.)}\\
  \multicolumn{2}{|c||}{}  & Mean & Std. & Mean & Std. & Mean & Std.\\
 \hline

 \multirow{4}{4em}{Cafe (Small)} & DEP (Ours) & \textbf{4.77} & 0.73 & \textbf{43.12} & 9.90 & \textbf{0.17} & 0.02\\
 & RH-NBV \cite{5} & 7.12 & 1.40 & 76.80 & 14.91 & 0.50 & 0.10\\
 & Frontier \cite{2} & 6.44 & 0.52 & 56.11 & 5.10 & 0.37 & 0.15\\
 & AEP \cite{7} & 5.48 & 0.73 & 59.16 & 7.87 & 0.37 & 0.15\\
 \hline

 \multirow{4}{4em}{Maze (Medium)} & DEP (Ours) & \textbf{17.75} & 0.71 & \textbf{146.44} & 21.19 & \textbf{1.05} & 0.11 \\
 & RH-NBV \cite{5} & 31.44 & 3.80 & 271.60 & 31.93 & 8.10 & 1.01 \\
 & Frontier \cite{2} & 34.74 & 3.35 & 330.10 & 31.13 & 2.01 & 1.15 \\
 & AEP \cite{7} & 23.23 & 2.23 & 200.65 & 19.29 & 2.44 & 1.03 \\
 \hline

 \multirow{4}{4em}{Office (Large)} & DEP (Ours) & \textbf{31.84} & 2.63 & 318.58 & 78.94 & \textbf{2.45} & 0.31 \\
 & RH-NBV \cite{5} & 48.21 & 9.89 & 421.80 & 80.92 & 12.13 & 2.83 \\
 & Frontier \cite{2} & 38.14 & 13.80 & \textbf{253.63} & 43.26 & 12.29 & 9.49\\
 & AEP \cite{7} & 37.46 & 2.47 & 327.78 & 32.59 & 6.64 & 2.47\\
 \hline

\end{tabular}
\label{exp_comparison}
\end{center}
\end{table*}

\section{Discussion}

\subsection{Exploration Performance Analysis}
In the Cafe (Small) and Maze (Medium), it is observed that the exploration rates of the frontier exploration and RH-NBV both have a significant drop. This drop can be explained by failing to generate the most informative path. RH-NBV is prone to generate \say{crooked} trajectories, increasing the motion time, and also fails to sample in high-utility regions with a limited sample number. The frontier planner does not determine the exploration goals regarding the exploration utility, leading to myopic behaviors and back-and-forth trajectories in the explored region. AEP uses the cached nodes as frontiers to guide the exploration when the information gain is low, resulting in a higher exploration rate, but still fails to evaluate the entire trajectory's utility. Since DEP incrementally builds the roadmap, each area's utility can be easily obtained to determine the high utility trajectory generation's optimal viewpoints. The trajectory optimization further improves the path quality to counter the nodes' sampling randomness. Especially in complex environments like Maze (Medium), where the viewpoint and trajectory utility become more critical in guiding the exploration, DEP performs much better. Unlike the previous two environments, the Office (Large) has larger open and free areas with multiple optimal viewpoints and trajectories. So, all the planners can obtain similar-quality trajectories and similar exploration rates. However, TABLE \ref{exp_comparison} shows that our planner's total exploration time is shorter than the others because that RH-NBV and frontier-based exploration spend a longer time to discover small rooms at the very end of the exploration.  Since the maximum sample number is limited in RH-NBV, it will take many iterations to find small unexplored rooms when the target area is large. From the computational-time perspective, our planner spends the least time for all three environments. Since our planner reuses the incrementally built roadmap, it can reduce unnecessary and repeated computation.  

\subsection{Evaluation of ESDF-based Optimization}
Fig. \ref{comparison_figure} left shows our planner's overall performance comparison with and without trajectory optimization. We compare the exploration time and path length in percentage by the average of ten experiments for each environment. The proposed optimization scheme is highly effective in improving the outcome of our planner. The trajectory-optimized planner's total exploration time is 77.01\% of the non-optimized one, and the total path length of the optimized one is 76.09\% of its counterpart. Specifically, TABLE \ref{optimization_comparison_table} gives the optimization statistics at each planning iteration across different environments. In each iteration, the optimized trajectory's average time is 87.28\%, and the path length is 91.86\% of the non-optimize one. The optimized trajectory is 20.58\% larger than the non-optimized trajectory for the nearest obstacle's average distance.  

\subsection{Evaluation of Replanning Time}
To quantitatively compare four planners' replanning time, we run experiments in the Dynamic Auditorium, Dynamic Tunnel, and Dynamic Field environments. The replanning time is recorded for each planner when the dynamic obstacles are encountered. The comparison of the replanning time with the benchmarks is shown in Fig. \ref{comparison_figure} right. The result shows that our planner has the least replanning time, and the RH-NBV and the frontier exploration take approximately twice the time as ours. Besides, AEP's replanning time is also 61.3\% longer than ours. The reduction of replanning time is due to the reuse of the previously maintained roadmap. 

\begin{figure}[t]
    \centering
    \includegraphics[scale=0.32]{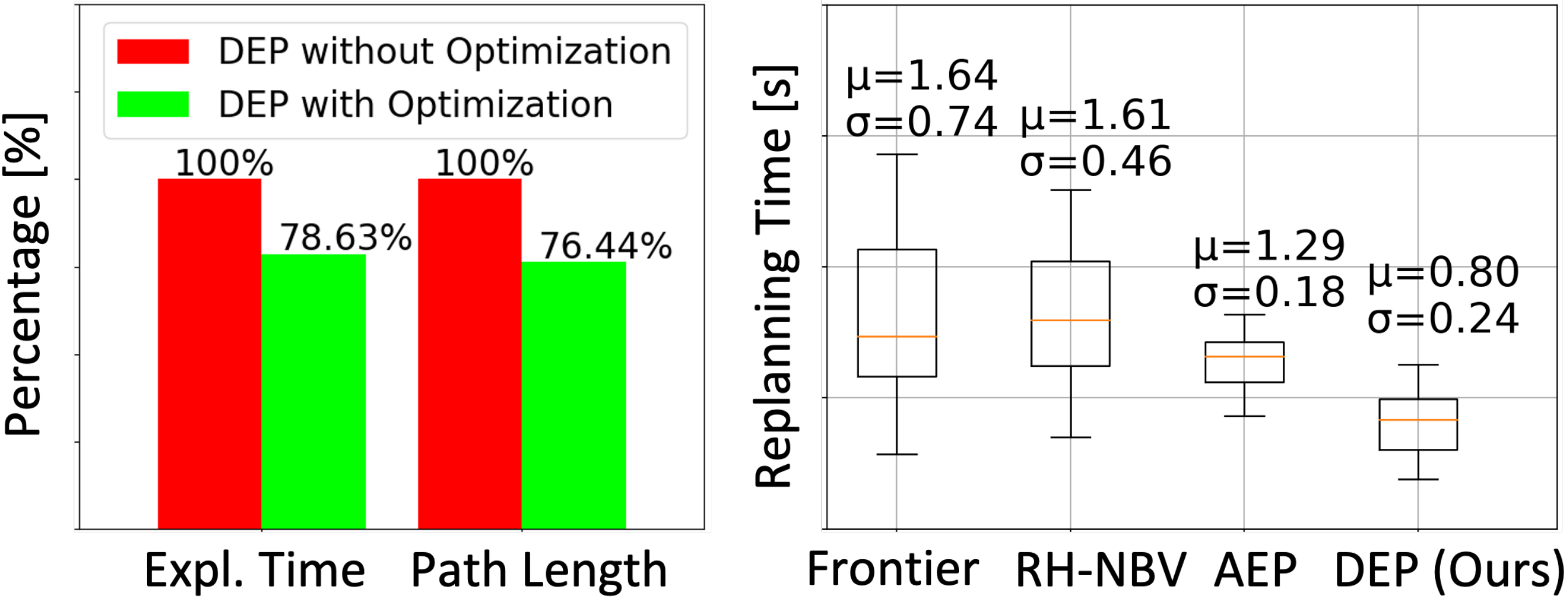}
    \caption{Left: Comparison of the overall exploration performance of our planner with and without optimization. Right: Comparison of the three planners' replanning time. }
    \label{comparison_figure}
\end{figure}

\section{Conclusion and Future Work}
In this paper, we present the novel dynamic exploration planner (DEP) for UAV exploration. Our algorithm extends the idea of incremental sampling by Probabilistic Roadmap (PRM). The proposed algorithm beats the benchmarks in exploration time, path length, and computational time. The evaluation indicates that our ESDF-based optimization further improves exploration performance. Besides, our planner shows the ability to explore dynamic environments safely. 

However, due to the robot sensor's limitation, dynamic obstacles cannot be completely detected. So, our assumption that the robot only needs to avoid its detected obstacle may not guarantee safety. A central detection system may be applied to track the dynamic obstacles in our future work.

\bibliographystyle{unsrt}
\bibliography{bibliography.bib}

\end{document}